\documentclass[runningheads]{llncs}
\usepackage{graphicx}
\usepackage{makecell}
\usepackage{multirow}
\usepackage{xcolor}
\usepackage{color}
\usepackage[caption=false]{subfig}

\usepackage{url}            
\usepackage{booktabs}       
\usepackage{amsfonts}       
\usepackage{nicefrac}       
\usepackage{microtype}      

\usepackage{todonotes}

\title{Self-Supervised Domain Adaptation for Diabetic Retinopathy Grading using \\ Vessel Image Reconstruction}
\titlerunning{Self-Supervised Domain Adaptation for Diabetic Retinopathy Grading}

\author{Duy M. H. Nguyen \inst{1} \and
Truong T. N. Mai\inst{2} \and
Ngoc T. T. Than \inst{3} \and
Alexander Prange \inst{1} \and
Daniel Sonntag \inst{1,4}\\}

\authorrunning{Duy M. H. Nguyen et al.}

\institute{German Research Center for Artificial Intelligence (DFKI)\\ Saarland Informatics Campus Saarbrücken, Germany\\ \and Department of Multimedia Engineering, Dongguk University, South Korea \and
Byers Eye Institute, Stanford University, United States \and 
Oldenburg University, Germany\\
}

\begin{document}

\maketitle

\begin{abstract}
This paper investigates the problem of domain adaptation for diabetic retinopathy (DR) grading. We learn invariant target-domain features by defining a novel self-supervised task based on retinal vessel image reconstructions, inspired by medical domain knowledge.Then, a benchmark of current state-of-the-art unsupervised domain adaptation methods on the DR problem is provided. It can be shown that our approach outperforms existing domain adaption strategies. Furthermore, when utilizing entire training data in the target domain, we are able to compete with several state-of-the-art approaches in final classification accuracy just by applying standard network architectures and using image-level labels.


\end{abstract}


\section{Introduction}
Diabetic retinopathy (DR) is a type of ocular disease that can cause blindness due to damaged blood vessels in the back of the eye. The causes of DR are high blood pressure and high blood sugar concentration, which are very common in modern lifestyles \cite{yun2008identification}. People with diabetes usually have higher risks of developing DR. In fact, one-third of diabetes patients show the symptoms of diabetic retinopathy according to recent studies \cite{zhou2020benchmark}. Therefore, early detection of DR is critical to ensure successful treatment. Unfortunately, detecting and grading diabetic retinopathy in practice is a laborious task, and DR is difficult to diagnose at an early stage even for professional ophthalmologists. As a result, developing a precise automatic DR diagnostic device is both necessary and advantageous.

Automated DR diagnosis systems take retinal images (fundus images) and yield DR grades. In the common retinal imaging dataset of DR, the grades of DR can be categorized into five stages \cite{gulshan2016development}: 0 - no DR, 1 - mild DR, 2 - moderate DR, 3 - severe DR, and 4 - proliferative DR. Specifically, the severity of DR is determined by taking the numbers, sizes, and appearances of lesions into account. For instance, figure 1 provides an illustration of five DR grades in the Kaggle DR dataset \cite{kaggle}. As can be seen, the characteristics of DR grades are complex in both structure and texture aspects. Therefore, automated diagnosis systems are required to be capable of extracting meaningful visual features from retinal images for precise DR grading. 

\begin{figure}
	\centering
	\subfloat[Grade 0]{%
		\includegraphics[width=0.18\textwidth]{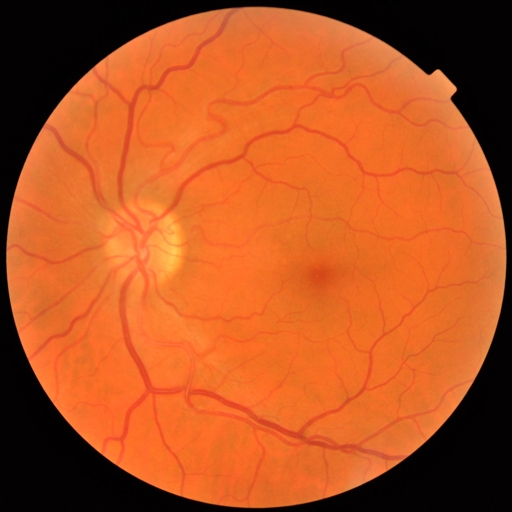}%
	}\,
	\subfloat[Grade 1]{%
		\includegraphics[width=0.18\textwidth]{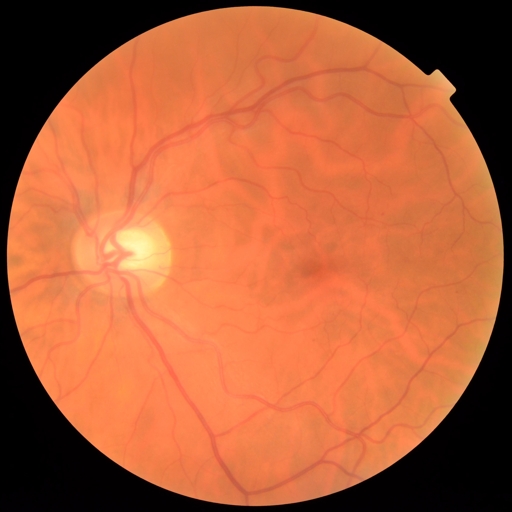}%
	}\,
	\subfloat[Grade 2]{%
		\includegraphics[width=0.18\textwidth]{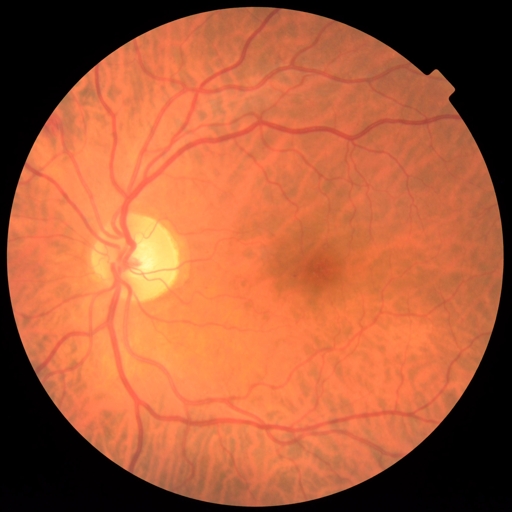}%
	}\,
	\subfloat[Grade 3]{%
		\includegraphics[width=0.18\textwidth]{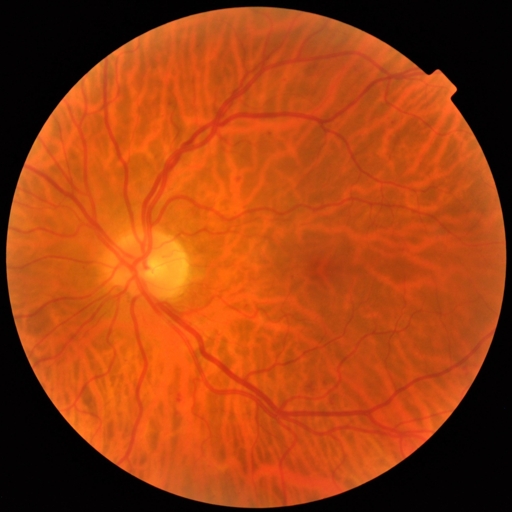}%
	}\,
	\subfloat[Grade 4]{%
		\includegraphics[width=0.18\textwidth]{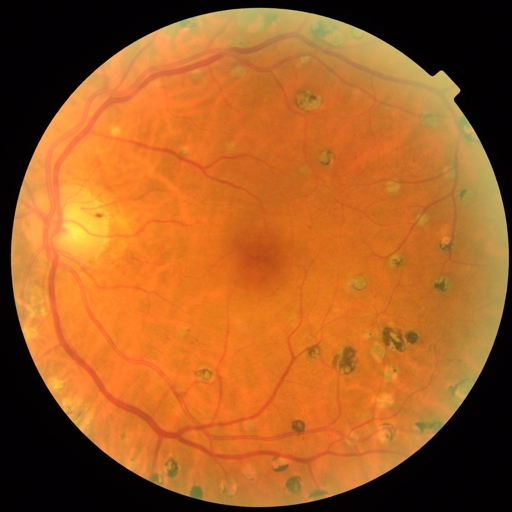}%
	}
	
	\caption{Illustration of different DR grades.}
	\label{fig:kaggle}
\end{figure}


With the success of deep learning, several CNN-based methods for DR grading of retinal images have been proposed. The paper from 2016 \cite{gulshan2016development} introduces the development and validation of a deep learning algorithm for detection of diabetic retinopathy---with high sensitivity and specificity when compared with manual grading by ophthalmologists for identifying diabetic retinopathy.  Jiang et al. \cite{jiang2019interpretable} also propose an ensemble of conventional deep learning methods to increase the predictive performance of automated DR grading. Lin et al. \cite{lin2018framework} 
in other direction introduce a joint model for lesion detection as well as DR identification, in which the DR is inferred from the fusion of original images and lesion information predicted by an attention-based network. Similarly, Zhou et al. in \cite{zhou2019collaborative} apply a two-step strategy: first produce a multi-lesion mask by using a semantic segmentation component, then the severity of DR is graded by exploiting the lesion mask. Recently, Wu et al. \cite{wu2021jcs} address the problem in a similar way, the classification is performed by employing pixel-level segmentation maps.

While recent works have demonstrated its effectiveness when trained and tested on a single dataset, they often suffer from the domain adaptation problem in practice. In particular, medical images in clinical applications are acquired from devices of different manufactures that vary in many aspects, including imaging modes, image processing algorithms, and hardware components. Therefore, the performance of a trained network from a particular source domain can dramatically decrease when applied to a different target domain. One possible way to overcome this barrier is to collect and label new samples in the target domain, which is necessary for fine-tune trained networks. Nevertheless, this task is laborious and expensive especially with medical images, as the data are limited and labeling requires extreme caution. As a result, it is highly desirable to develop an algorithm that can adapt well in the new domain without additional labeled data for training. Such an approach is known as unsupervised domain adaption.

In this paper, we propose a self-supervised method to reduce domain shift in the fundus images' distribution by learning the invariant feature representations. To this end, feature extraction layers are trained by using both labeled data from the source domain and a self-supervised task on a target domain by defining image reconstruction tasks around retinal vessel positions. Moreover, we also incorporate additional restricted loss functions throughout the training phase to encourage the acquired features to be consistent with the main objective.  

At a glance, we make three main contributions. First, we address the domain adaptation problem for DR grading on fundus images using a novel self-supervised approach motivated by medical domain knowledge. Second, we provide a benchmark of current state-of-the-art unsupervised domain adaptation methods on the DR problem. Finally, we show that our approach when using fully training data in the target domain  obtains competitive performance just by employing standard network architectures and using image-level labels.

\begin{figure}
    \centering
    \includegraphics[width=1.0\textwidth]{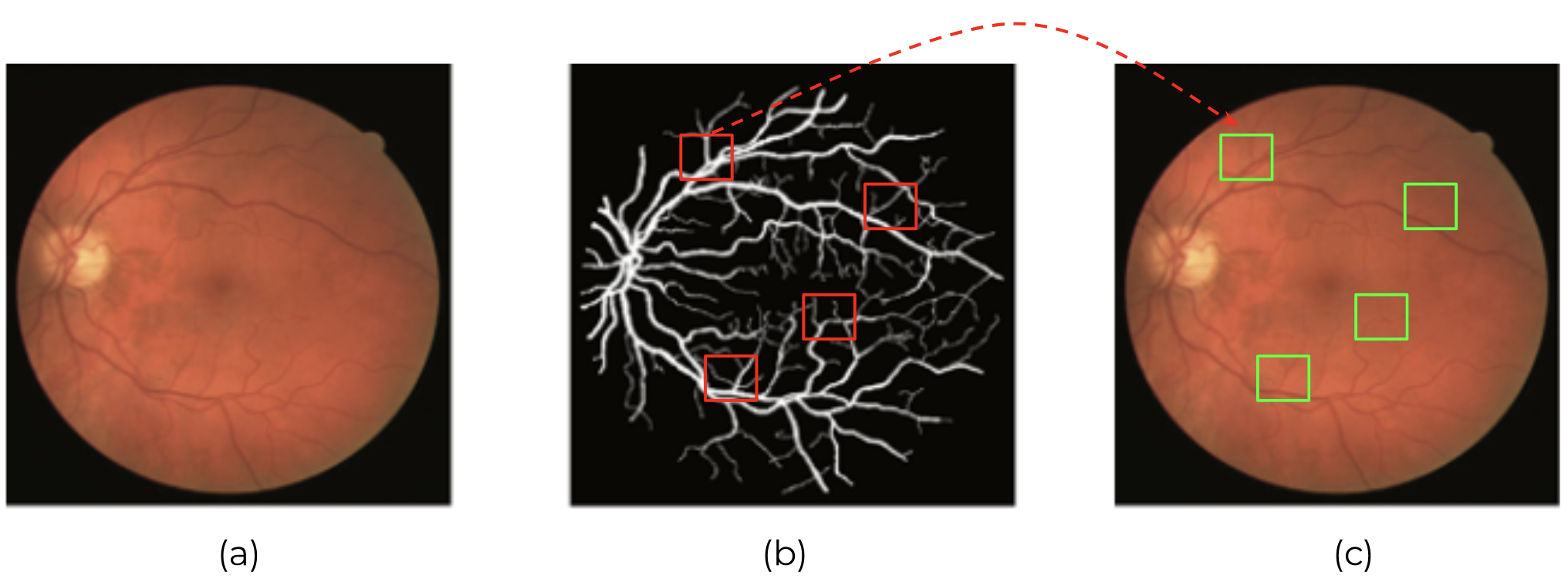}
    \caption{Illustration of our vessel segmentation reconstruction-based SSL. (a) input image $x^{t}$, (b) its vessel segmentation $y_{v}^{t}$, (c) binary masks $B^{t}$ (inside green rectangles) sampled along edges of $y_{v}^{t}$ in (b) (red rectangles). The image regions inside $B^{t}$ are removed to define $\hat{y}^{t}$ and asking the encoder-decoder network to reconstruct them given the remaining pixels in $\hat{x}^{t}$.}
    \label{fig:SSL}
\end{figure}
\vspace{-0.3in}
\section{Related Work}
Over the last decade, research in domain adaption has achieved remarkable results. Tzeng et al. \cite{tzeng2014deep} propose a deep domain confusion technique to minimize the maximum mean discrepancy, a non-parametric metric of distribution divergence proposed by Gretton et al. \cite{gretton2012kernel}, so that the divergence between two distributions is reduced. The algorithm developed by Sun et al. \cite{sun2016deep} is an extension of their previous work \cite{sun2016return}, in which CNNs are employed to learn a nonlinear transformation for correlation alignment. Recently, Wang et al. \cite{wang2020domain} have presented a domain adaptation algorithm for screening normal and abnormal retinopathy in optical coherence tomography
(OCT) images. The system consists of several complex components guided by the Wasserstein distance \cite{shen2018wasserstein} to extract invariant representations across different domains.

In other directions, researchers have employed generative adversarial networks (GANs) to learn better invariant features. Tzeng et al. \cite{tzeng2017adversarial} combine discriminative modeling with untied weight sharing and a GAN-based loss to create an adversarial discriminative domain algorithm. Shen et al.'s algorithm \cite{shen2018wasserstein} extracts domain invariant feature representations by optimizing the feature
extractor network, which minimizes the Wasserstein distance trained in an adversarial manner between the source and target domains. In a different way, Long et al.~\cite{long2017conditional} design a conditional domain adversarial network by exploiting two strategies, namely multilinear conditioning, to capture the cross-domain covariance, and entropy conditioning, to ensure the transferability. 

Our method in this paper follows the self-supervised learning (SSL) approach~\cite{kolesnikov2019revisiting}, which is recently an active research direction due to its effectiveness in learning feature representations. In particular, SSL refers to a representation learning method where a supervised task is defined over unlabelled data to reduce the data labeling cost and leverage the available unlabelled data. Until now, several algorithms based on SSL have been introduced. The method presented by Xu et al. \cite{xu2019self} is a generic network with several kinds of learning tasks in SSL that can adapt to various datasets and diverse applications. In medical image analysis, authors in \cite{chen2019self} introduce a SSL pretext task based on context restoration, thereby two isolated small regions are selected randomly and swap their positions. A deep network is then trained to recover original orders in input images. Unfortunately, these prior works are mostly designed in the same domain. Recently, Xiao et al.~\cite{xiao2021self} have pioneered to apply the SSL method for domain adaptation problems. Specifically, target-domain-aware features are learned from unlabeled data for image classification through an image rotation-based pretext task trained by a unified encoder for both source and target domains.

\textbf{Difference  w.r.t.  Previous  Work:}
Our method follows Xiao
et al. \cite{xiao2021self}; however, we make the following modifications for our setting. First, rather than a rotation task like \cite{xiao2021self}, we study medical domain knowledge to create a novel SSL prediction task, i.e., vessel segmentation reconstruction that has a solid connection to the severity of diabetic retinopathy \cite{gulshan2016development,haneda2010international}. Second, a two-player procedure is integrated through a discriminate network to ensure mission regions generated in SSL tasks look realistic and consistent with the image context. As a results, our objective function has more constraints on learned features when compared to \cite{xiao2021self}.   

\section{Method}
\begin{figure}[!hbt]
    \centering
    \includegraphics[width=0.8\textwidth]{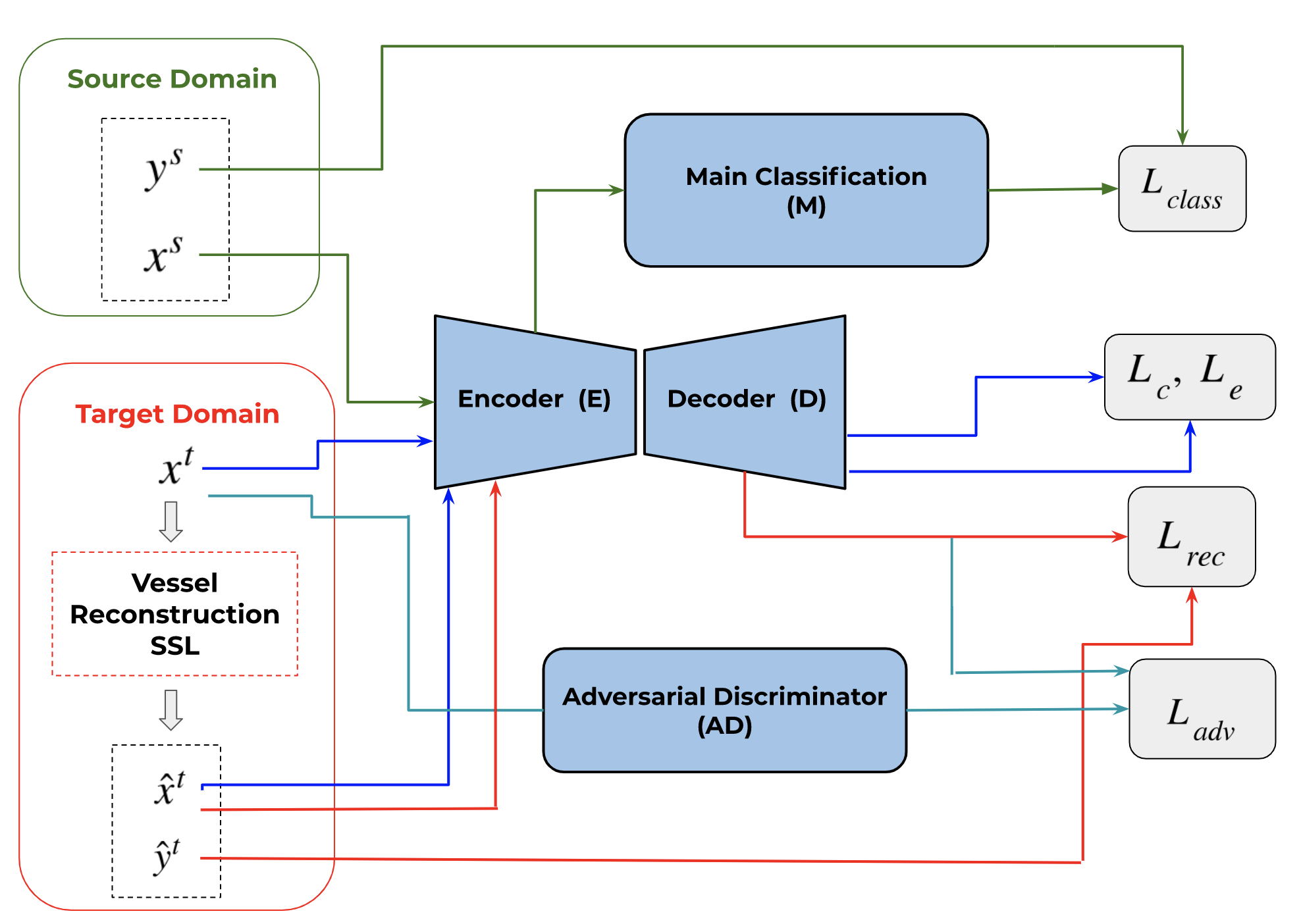}
    \caption{Overview our proposed unsupervised domain adaption with vessel reconstruction-based self-supervised learning.}
    \label{fig:overview}
\end{figure}
\label{sec:methods}
\subsection{Overview}
Our proposed method aims at learning invariant features across different domains through encoder layers shared to optimize several relevant tasks. In specific, we define labeled images in the source domain $X_s = \{(x_{i}^{s},\, y_{i}^{s})\}_{i=0}^{N_s}$ with $y_{i}^{s}$ is the corresponding label (DR grades) of image $x_{i}^{s}$ and $N_{s}$ is the total of images. In the target domain, we assume that only a set of unlabeled images denoted by $X_t = \{x_{i}^{t}\}_{i=0}^{N_t}$ with $N_t$ samples is available. Our framework, which uses labeled $X_s$ and unlabeled $X_t$ for domain adaptation, consists of four distinct blocks: an encoder network $E$, a decoder network $D$, an adversarial discriminator $AD$, and a main classifier $M$. These blocks are parameterized by $\theta_{e},\, \theta_{d}, \theta_{ad}$ and $\theta_{m}$ respectively. For each image $x_{i}^{t} \in X_{t}$, we transform it through the self-supervised learning task based on vessel image reconstruction to define a new set $\hat{X_t} = \{(\hat{x}_{i}^{t}, \hat{y}_{i}^{t})\}_{i=1}^{\hat{N}_{t}}$, which are used to train $E$ and $D$ blocks for predicting removed sub-patch images. To encourage that the reconstructed regions look authentic, the adversarial discriminator $AD$ is integrated through the two-player game learning procedure for distinguishing generated and ground-truth samples. Finally, the block $M$ is built on top of the encoder layer $E$ and acts as the main classification task. We describe below each aforementioned architecture in detail.

\subsection{Retinal Vessel Reconstruction-based SSL} 
According to medical protocol \cite{gulshan2016development,haneda2010international}, the severity of
DR can be predicted by observing the number and size of related lesion appearances and complications. While their positions tend to cluster near vessel positions, we use this attribute to create a new SSL task that forces learnt feature representation to capture such lesions.

Given a sample $x^{t} \in X_{t}$, we extract its vessel segmentation image $y_{v}^{t} = f(x_{i}^{t})$ with $f(.)$ is a trained deep network (Figure \ref{fig:SSL}a and \ref{fig:SSL}b). In this work, we use $f(.)$ as a proposed architecture in \cite{sun2020robust}. Let $B^{t}$ is a binary mask corresponding to the dropped image region in $x^t$, with a value of $1$ if a pixel was dropped and $0$ for input pixels. Unlike related works \cite{chen2019self,doersch2015unsupervised}, we generate region masks in $B^t$ by randomly sampling sub-patch images along vessel positions in $y_{v}^{t}$ as indicated in Figure \ref{fig:SSL}c. We then define a new pair of samples:
\begin{equation}
    \hat{x}^{t} = (1-B^t) \odot x^{t},\, \hat{y}^{t} = B^t \odot x^{t}
\end{equation}
where $\odot$ is the element-wise product operation.

\paragraph{\textbf{Reconstruction Loss}}
We train a context encoder $F$ formed from the encoder $E$ and the decoder $D$ to reconstruct target regions $\hat{y}^{t}$ given the input $\hat{x}^{t}$. A normalized $L2$ distance is employed as our reconstruction objective function:

\begin{equation}
    L_{rec} = \min_{\theta_{e},\, \theta_{d}}  \mathbb{E}_{x^{t} \in X^{t}}||B^t \odot F(\hat{x}^{t}) - \hat{y}^{t}||_{2}^{2}
\end{equation}
\paragraph{\textbf{Adversarial Loss}}
The objective function $L_{rec}$ takes into account the overall construction of the missing region and agreement with its context, but tends to average together the multiple forms in predictions. We thus adapt the adversarial discriminator $AD$ as \cite{pathak2016context,wang2020attentive} to make the predictions of the context encoder $F$ look real through selecting similar instances from the target distribution. The joint min-max objective function of discriminator $AD$ and generator $F$ is:
\begin{equation}
    L_{adv} = \min_{\theta_{e},\, \theta_{d}}\max_{\theta_{ad}} \mathbb{E}_{x^{t} \in X^{t}}[\log(D(x^{t}) + \log(1 - D(F(\hat{x}^{t})))]
\end{equation}
By jointly optimizing $L_{rec}$ and $L_{adv}$, we encourage the output of the context encoder $F$ to look realistic on the entire prediction, not just the missing regions as in $L_{rec}$.

\subsection{Relevant Features from SSL}
\paragraph{\textbf{Main Classification Loss}}
In our framework, the block $M$ takes the feature representation from the encoder $E$ to predict a corresponding label $y^{t}$ for each image $x^t$ in the target domain. The network $M$ and encoder $E$ are trained with labeled data in the source domain by optimizing the classification problem:
\begin{equation}
    L_{class} = \min_{\theta_{e},\theta_{m}} \mathbb{E}_{x^{s},y^{s}\in X^{s}}[-\log p(y^{s}|x^{s})]
    \label{eq:l_class}
\end{equation}
where $p(y^{s}|x^{s})$ is a conditional probability distribution of $y^{s}$ given $x^{s}$ parameterized by $E$ and $M$ networks.

\paragraph{\textbf{Constrained Features from SSL}}
While the SSL task is designed to encourage the encoder $E$ to capture invariant features across different domains and pay attention to vessel positions, there is no guarantee of the compatibility between this SSL task and the main classification target. Inspired from prior works in semi-supervised learning \cite{miyato2018virtual,NEURIPS2020_44feb009}, we adapt two additional loss constraints on the feature representation generated by the SSL $\hat{x}^{t}$, the input $x^{t}$, and the target label $y^{t}$:
\begin{equation}
    L_{c} = \min_{\theta_{e},\theta_{m}} \mathbb{E}_{x^{t} \in X^{t}}\mathbb{E}_{\hat{x}^{t} \in \hat{X}^{t}}\big[D_{KL}(\hat{p}(y^{t}|x^{t})||p(y^{t}|\hat{x}^{t}))\big]
\end{equation}
\vspace{-0.1in}
\begin{equation}
    L_{e} = \min_{\theta_{e},\theta_{m}}\mathbb{E}_{x^{t} \in X^{t}}\big[-\sum_{y^{t}} p(y^{t}|x^{t})\log(p(y^{t}|x^{t}))\big]
\end{equation}
where $D_{KL}$ is the Kullback-Leibler consistency loss \cite{miyato2018virtual,NEURIPS2020_44feb009}, $\hat{p}(y^{t}|x^{t})$ is a fixed copy
of the current $p(y^{t}|x^{t})$ with parameters $\theta_{e},\theta_{m}$, it means that $\hat{p}(y^{t}|x^{t})$ is only used for each inference step and the gradient is not propagated through them.

Intuitively, the consistency objective function $L_c$ forces the feature representation in $E$ to be insensitive to data augmentation in defined SSL task while the objective $L_e$ penalizes uncertain predictions, leading to more discriminative representations. However, the equations $L_c,\, L_e$ require labels $y^{t}$ in the target domain to optimize, which are assumed to be not available in our unsupervised domain adaption. We address this challenge by integrating pseudo-labels $y^{t}$ generated by predictions using $E$ and $M$ blocks and updating it progressively after each training step.

\paragraph{\textbf{Overall Objective Function}}
In summary, our overall objective function is:
\begin{equation}
    L = L_{class} + \lambda_{rec}L_{rec} + \lambda_{adv}L_{adv} + \lambda_{c}L_{c} + \lambda_{e}L_{e}
\end{equation}
where $\lambda_{rec},\, \lambda_{adv},\, \lambda_{c},\, \lambda_{e}$ are coefficients of corresponding objective functions. Due to the generative adversarial function in $L_{adv}$, $L$ is the min-max objective problem. We adapt the alternative optimization strategy to first update parameters $\theta_{e}, \theta_{d}, \theta_{m}$, second update $\theta_{ad}$ and repeating this process until convergence. In our experiment, we use feature extraction layers from ResNet-50 \cite{he2016deep} for both the encoder $E$, decoder $D$ and adversarial discriminator $AD$. These layers are shaped in certain architectural constraints as in \cite{radford2015unsupervised}. For
the main classification $M$, we adapt a simple average pooling followed by a fully connected layer.

\section{Experiments and Results}
\label{sec:experiment}

\subsection{Evaluation Method}
We assess our method, denoted as VesRec-SSL, in two DR grading scenarios: unsupervised domain adaption (UDA) and conventional classification problems. In the first case, all UDA methods are trained using both supervised samples in the source domain and unlabeled samples in the target domain. The performance is then evaluated using the target domain's testing set.  In the second case, we train and test in the same domain, i.e., the training set's labeled images are utilized in the training step, and trained networks are measured on the remaining data. For the latter case, our method may be viewed as a pre-training phase~\cite{chen2019self,nguyen2021tatl}; thereby, obtained weights after training VesRec-SSL will be used in the fine-tuning step using partially or completely supervised training samples. 

\subsection{Dataset and Metrics}
We employ two DR-graded retinal image datasets, Kaggle EyePACS \cite{kaggle} and FGADR \cite{zhou2020benchmark}, for training and testing with DR gradings from 0-4 (Figure \ref{fig:kaggle}). We follow the splitting standard in EyePACS with $35126$ training images and $53576$ testing images. With the FGADR dataset, we can only access $1842$ images (SegSet) out of a total of $2842$ images at the moment due to data privacy. Because there is no specific train/test on the SegSet, we apply 3-fold cross-validation to compute the final performance. For quantitative metrics, we use classification accuracy and Quadratic Weighted Kappa (Q. W. Kappa) \cite{zhou2020benchmark}. 

\subsection{Performance of Unsupervised Adaption Methods}
In this task, we choose one dataset as the source domain and the other as the target domain. We provide a benchmark of three different methods in literature: Xiao et al.'s Rotation-based SSL \cite{xiao2021self}, Long et al.'s CDAN and CDAN-E \cite{long2017conditional}. For fairly comparison, we choose ResNet50 as the backbone network for all methods. The quantitative evaluation is shown in Table \ref{tab:table-1} where ``EyePACS $\rightarrow$ FGADR (SegSet)'' indicates the source domain is EyePACS the target domain is FGADR restricted on SegSet with $1842$ images, and similarly for ``FGADR (SegSet) $\rightarrow$ EyePACS''. In practice, we found that training baselines directly in our setting is not straightforward due to the imbalance among grading types and the complexity of distinguishing distinct diseases. Therefore, we applied the following training methods: 
\begin{itemize}
\item First, we only activate the main classification loss using fully supervised samples in the source domain in the initial phase and training until the model converges. Next, auxiliary loss functions will be activated, and the network is continued to train in the latter phase.
\item  Second, we apply the progressive resizing technique introduced in the fast.ai\footnote[1]{https://course.fast.ai/}, and the DAWNBench challenge \cite{coleman2017dawnbench} in which the network is trained with smaller images at the beginning, and obtained weights are utilized for training another model with larger images. We use two different resolutions in our setting:  $256 \times 256$ and  $512 \times 512$.
\item Finally, the optimal learning rate is automatically chosen by the Cyclical Learning method \cite{smith2017cyclical} with the SGD optimizer \cite{goodfellow2016deep}, which sets the learning rate to cyclically change between reasonable boundary values. 

\end{itemize}

As shown in Table 1, our VesRec-SSL outperforms competitors by a remarkable margin in all settings and metrics. For instance, we achieve $2-3\%$ more for FGADR and $1-3\%$ more for EyePACS, compared to the second competitor CDAN-E. In addition, we can observe that the performance in ``FGADR (SegSet) $\rightarrow$ EyePACS'' is lower than that in ``EyePACS $\rightarrow$ FGADR (SegSet)'' in most of the cases. We argue this 
happens due to the number of training instances in the source domain of ``FGADR``, which is much lower than that of ``EyePACS``. 


\begin{table*}[t]
\centering
\caption{Performance of Unsupervised Domain Adaption Methods.}
\scalebox{0.95}{
\begin{tabular}{c|l|l|l|l|l|l}
\Xhline{2\arrayrulewidth}
\multirow{2}{*}{Method}                  & \multicolumn{3}{c|}{\small{EyePACS $\rightarrow$ FGADR (SegSet)}}                 & \multicolumn{3}{c}{\small{FGADR (SegSet) $\rightarrow$ EyePACS}}                 \\ \cline{2-7} 
                                         & \multicolumn{2}{c|}{\hspace{0.1in}Acc.\,\hspace{0.1in}\ } & \hspace{0.1in}Q.W. Kappa\,\hspace{0.1in}               & \multicolumn{2}{c|}{\hspace{0.1in} Acc.\,\hspace{0.1in}\ } & \hspace{0.1in}Q.W. Kappa\hspace{0.1in}               \\ \hline
\multicolumn{1}{l|}{Rotation-based SSL \cite{xiao2021self}} & \multicolumn{2}{c|}{0.728}     &    \multicolumn{1}{c|}{0.672}                      & \multicolumn{2}{c|}{0.681}     &    \multicolumn{1}{c}{0.660}                      \\ \hline
CDAN     \cite{long2017conditional}                             & \multicolumn{2}{c|}{0.741}     &     \multicolumn{1}{c|}{0.685}                     & \multicolumn{2}{c|}{0.697}     &    \multicolumn{1}{c}{0.685}                        \\ \hline
CDAN-E   \cite{long2017conditional}                                 & \multicolumn{2}{c|}{0.755}     &    \multicolumn{1}{c|}{0.706}                     & \multicolumn{2}{c|}{0.702}     &        \multicolumn{1}{c}{0.691}                  \\ \hline
VesRec-SSL (Our) & \multicolumn{2}{c|}{\textbf{0.782}}     & \multicolumn{1}{c|}{\textbf{0.725}}    & \multicolumn{2}{c|}{\textbf{0.736}}     & \multicolumn{1}{c}{\textbf{0.702}}    \\ \Xhline{2\arrayrulewidth}
\end{tabular}}
\label{tab:table-1}
\end{table*}

\begin{table*}[!h]
\centering
\caption{Performance of competitor methods on the DR grading prediction. Red, blue, black, and orange represent the top four best results.}
\begin{tabular}{c|l|l|l}
\Xhline{2\arrayrulewidth}
\multirow{2}{*}{Method}                  & \multicolumn{3}{c}{EyePACS}             \\ \cline{2-4} 
                                         & \multicolumn{2}{c|}{\hspace{0.1in}Acc.\,\hspace{0.1in}\ } & \hspace{0.1in}Q.W. Kappa\,\hspace{0.1in} \\ \hline
\multicolumn{1}{c|}{VGG-16 \cite{simonyan2014very}} & \multicolumn{2}{c|}{0.836}     &    \multicolumn{1}{c}{0.820}    \\ 
ResNet-50 \cite{he2016deep}               & \multicolumn{2}{c|}{0.846}     &     \multicolumn{1}{c}{0.824} \\ 
Inception v3 \cite{szegedy2016rethinking}             & \multicolumn{2}{c|}{0.840}     &    \multicolumn{1}{c}{0.811}  \\ 
DenseNet-121 \cite{huang2017densely}                      & \multicolumn{2}{c|}{0.854}     & \multicolumn{1}{c}{0.835}  \\ 
Lin et al., \cite{lin2018framework}                                    & \multicolumn{2}{c|}{0.867}     & \multicolumn{1}{c}{0.857}\\ \hline
Zhou et al., \cite{zhou2019collaborative}                                   & \multicolumn{2}{c|}{\textcolor{red}{0.895}}     & \multicolumn{1}{c}{\textcolor{red}{0.885}}\\ 
Wu et al., \cite{wu2021jcs}                                   & \multicolumn{2}{c|}{\textcolor{blue}{0.886}}     & \multicolumn{1}{c}{\textcolor{blue}{0.877}}  \\ \hline
VesRec-SSL (ResNet-50) + $0\%$                                   & \multicolumn{2}{c|}{0.736}     & \multicolumn{1}{c}{0.702}  \\ 
VesRec-SSL (ResNet-50) + $50\%$                                   & \multicolumn{2}{c|}{0.798}     & \multicolumn{1}{c}{0.774}    \\ 
VesRec-SSL (ResNet-50) + $100\%$                                   & \multicolumn{2}{c|}{0.864}     & \multicolumn{1}{c}{0.852}  \\ \hline
VesRec-SSL (DenseNet-121) + $0\%$                                   & \multicolumn{2}{c|}{0.744}     & \multicolumn{1}{c}{0.711}  \\ 
VesRec-SSL (DenseNet-121) + $50\%$                                   & \multicolumn{2}{c|}{0.815}     & \multicolumn{1}{c}{0.793}    \\ 
VesRec-SSL (DenseNet-121) + $100\%$                                   & \multicolumn{2}{c|}{\textcolor{orange}{0.871}}     & \multicolumn{1}{c}{\textcolor{orange}{0.862}}  \\ \hline
VesRec-SSL (ResNet-50 + DenseNet-121) + $100\%$                                   & \multicolumn{2}{c|}{\textbf{0.891}}     & \multicolumn{1}{c}{\textbf{0.879}}  \\ 
\Xhline{2\arrayrulewidth}
\end{tabular}
\label{tab:table-2}
\end{table*}

\subsection{Performance of Baseline Methods on DR Grading Prediction}
In this task, we compare our algorithm to the most recent state-of-the-art method reported in \cite{zhou2020benchmark}. Due to the data privacy on the FGADR dataset, we can only benchmark baselines on the EyePACS dataset. For ablation studies, we also fine-tune our VesRec-SSL with additional 0\%, 50\%, and 100\% labeled data pairs from the target domain. The evaluation results are shown in Table \ref{tab:table-2}. Besides the default backbone with ResNet-50, we consider a variation with DenseNet-121 network for fairly evaluation with two top methods in \cite{zhou2019collaborative,wu2021jcs}. Moreover, we also utilize feature extraction layers as average pooling of feature maps obtained from ResNet-50 and DenseNet-121 at the last row and train this network with $100\%$ training data.

The results indicate that without labeled data from the target domain, our two settings perform considerably worse than all baselines trained with fully supervised images. However, by progressively increasing the amount of labeled data from $50\%$ to $100\%$, we can significantly increase performance.
For example, the ResNet-50 with $50\%$ data outperforms the $0\%$ case with approximately $6/7\%$ in Acc/Q.W.Kappa. DenseNet-121 follows a similar pattern, improving $7/8\%$ ($50\%$ data), and even with $100\%$ training data, our VesRec-SSL can achieve the fourth rank in total. Finally, we observe that utilizing both the ResNet-50 and DenseNet-121 backbones can result in a second-rank overall without modifying the network architecture or adding extra pixel-level segmentation maps for relative lesion characteristics as in \cite{wu2021jcs,zhou2019collaborative}. In summary, we argue that our method with vessel reconstruction-based SSL has proven effective for domain adaptation under DR grading applications, especially as partial or complete annotations are available.


\section{Conclusion}
Domain shift is a big obstacle of deep learning-based medical analysis, especially as images are collected by using various devices. In this work, we showed that the unsupervised domain adaption for diabetic retinopathy grading can benefit from our novel self-supervised learning (SSL) based on the medical vessel image reconstruction tasks. Furthermore, when fully integrating annotation data and simply using standard network architectures, our technique achieves comparable performance to cutting-edge benchmarks. In future work, we consider to extend the SSL task to include related lesion appearances such as microaneurysms (MAs), Hard exudates, and Soft exudates \cite{zhou2020benchmark} to acquire improved invariant feature representation guided by medical domain knowledge. Moreover, making our network's predictions understandable and explainable to clinicians is also a crucial question for further investigation based on our recent medical application projects \cite{nguyen2021attention,nguyen2020visually,nunnari2021software,sonntag2020skincare}. We also aim to investigate in the direction of information fusion and explainable AI by incorporating multimodal embeddings with Graph Neural Networks~\cite{holzinger2021towards,yuan2020explainability}.

\section*{Acknowledgement}
This research has been supported by the
Ophthalmo-AI project (BMBF, 16SV8639), the Ki-Para-Mi project (BMBF, 01IS19038B), the pAItient project (BMG, 2520DAT0P2), and the Endowed Chair of Applied Artificial Intelligence, Oldenburg University. We would like to thank all student assistants that contributed to the development of the platform, see iml.dfki.de.

%
%

%
%





\bibliographystyle{splncs04}  
\bibliography{references}



\end{document}